\documentclass[sigconf]{aamas} 
\usepackage{balance} % for balancing columns on the final page
\usepackage{subcaption}
\usepackage{algorithm2e}
\RestyleAlgo{ruled}  % puts border around algorithm heading
\LinesNumbered
\setcopyright{ifaamas}
\acmConference[AAMAS '23]{Proc.\@ of the 22nd International Conference
on Autonomous Agents and Multiagent Systems (AAMAS 2023)}{May 29 -- June 2, 2023}
{London, United Kingdom}{A.~Ricci, W.~Yeoh, N.~Agmon, B.~An (eds.)}
\copyrightyear{2023}
\acmYear{2023}
\acmDOI{}
\acmPrice{}
\acmISBN{}

\acmSubmissionID{???}

\title{Causal Counterfactuals for Improving the Robustness of Reinforcement Learning}

\author{Tom He}
\affiliation{
  \institution{Trinity College Dublin}
  \city{Dublin}
  \country{Ireland}}
\email{heto@tcd.ie}

\author{Jasmina Gajcin}
\affiliation{
  \institution{Trinity College Dublin}
  \city{Dublin}
  \country{Ireland}}
\email{gajcinj@tcd.ie}

\author{Ivana Dusparic}
\affiliation{
  \institution{Trinity College Dublin}
  \city{Dublin}
  \country{Ireland}}
\email{ivana.dusparic@tcd.ie}

\begin{abstract}
Reinforcement learning (RL) is used in various robotic applications. RL enables agents to learn tasks autonomously by interacting with the environment. The more critical the tasks are, the higher the demand for the robustness of the RL systems. Causal RL combines RL and causal inference to make RL more robust. Causal RL agents use a causal representation to capture the invariant causal mechanisms that can be transferred from one task to another. Currently, there is limited research in Causal RL, and existing solutions are usually not complete or feasible for real-world applications. In this work, we propose CausalCF, the first complete Causal RL solution incorporating ideas from Causal Curiosity and CoPhy. Causal Curiosity provides an approach for using interventions, and CoPhy is modified to enable the RL agent to perform counterfactuals. Causal Curiosity has been applied to robotic grasping and manipulation tasks in CausalWorld. CausalWorld provides a realistic simulation environment based on the TriFinger robot. We apply CausalCF to complex robotic tasks and show that it improves the RL agent's robustness using CausalWorld.
\end{abstract}


\newcommand{\BibTeX}{\rm B\kern-.05em{\sc i\kern-.025em b}\kern-.08em\TeX}

\begin{document}

%%% The following commands remove the headers in your paper. For final 
%%% papers, these will be inserted during the pagination process.

\pagestyle{fancy}
\fancyhead{}

%%% The next command prints the information defined in the preamble.

\maketitle 

%%%%%%%%%%%%%%%%%%%%%%%%%%%%%%%%%%%%%%%%%%%%%%%%%%%%%%%%%%%%%%%%%%%%%%%%

\section{Introduction}

Intelligent robots are increasingly being considered for use alongside humans in critical applications such as healthcare and transportation \cite{10.1145/3477600,9146378}. Reinforcement learning (RL) can enable agents or robots to learn tasks unsupervised \cite{sutton2018reinforcement}. RL agents learn tasks from scratch through an iterative exploration and exploitation process. Recent improvements in deep learning (DL) methods enabled deep RL (DRL), which uses deep neural networks (DNN) \cite{Goodfellow-et-al-2016} to represent policies and value functions because DNNs scale efficiently with the amount of data. Applications of RL in the real world, especially in critical applications, require RL agents to be trustworthy. The robustness and explainability of RL need to be improved for RL to be trustworthy in real-world deployments \cite{Puiutta2020ExplainableRL}.

Causal RL \cite{CausalRLElias} combines RL with causal inference \cite{pearl2000models,halpern2020causes} to make RL agents more robust and explainable. Environments are represented as structural causal models (SCM) \cite{pearl2000models} in Causal RL. SCMs are directed acyclic graphs, where nodes represent causal factors, and edges indicate causal relationships. The causal mechanisms remain invariant across tasks and can improve the robustness of RL \cite{Scholkopf2021TowardCR}. The RL agent learns about the underlying SCM through three different interactions: association (seeing), interventions (doing), and counterfactuals (imagining). All three interactions are needed to learn the \emph{complete} underlying SCM \cite{bar:etal2020,CausalRLElias}. The causal information obtained through interacting with the environment is then stored as a causal representation.

Currently, there is limited research on Causal RL. Existing approaches have limitations, such as assuming the SCM is known upfront, or partially known \cite{Ke2019LearningNC,Madumal2020ExplainableRL,dazeley2021explainable}, they are usually not complete  \cite{Sontakke2021CausalCR,Ahmed2021CausalWorldAR,Jin2020OfflineLO, Zhu2020CausalDW}, and they have not been applied to complex tasks like robotics \cite{Dasgupta2019CausalRF,Baradel2020COPHYCL}. We define a Causal RL solution to be \emph{complete} if the solution uses all three of the interactions and a causal representation \cite{CausalRLElias}. In this work, we propose Causal Counterfactuals (CausalCF), the first \emph{complete} Causal RL solution applied to complex robotic tasks. CausalCF learns about the underlying SCM from scratch using the three different Causal RL interactions. The causal knowledge is stored as an abstract causal representation in the form of a vector and can be concatenated to the states of the agent. The representation is scalable and transferable to different tasks. We implemented and evaluated CausalCF in a realistic robotic simulation environment called CausalWorld \cite{Ahmed2021CausalWorldAR} which provides a range of complex robotic tasks.

CausalCF incorporates ideas from Causal Curiosity \cite{Sontakke2021CausalCR} and CoPhy \cite{Baradel2020COPHYCL}. Causal Curiosity provides an approach for using interventions and a causal representation to train an RL agent, previously successfully applied in CausalWorld \cite{Ahmed2021CausalWorldAR}. In addition to interventions, CausalCF makes use of counterfactuals. CoPhy is a deep learning solution that can perform counterfactuals. In this work, we adapt the deep learning architecture from CoPhy, originally aimed at supervised learning only, for use in RL. The implementation of our proposed approach, CausalCF, is available on \url{https://github.com/Tom1042roboai/CausalCF}. The main contributions of this work are:
\begin{enumerate}
    \item We present CausalCF, the first complete Causal RL solution applied to robotic tasks.
    \item We adapt the CoPhy architecture from supervised learning to RL tasks, and incorporate it into CausalCF.
    \item We evaluate all the components of the CausalCF design in CausalWorld and show that they improve the training performance and the robustness of the RL agent.
    \item We evaluate the transferability of the causal representation, and confirm that it captures the invariant causal mechanisms across tasks.
\end{enumerate}

The rest of the paper is structured as follows. Section \ref{sec:background} introduces causal inference, reviews related work, and discusses the limitations of Causal Curiosity and CoPhy. Section \ref{sec:causalworld} presents the evaluation environment, CausalWorld. Section \ref{sec:causalCF} describes the design of our proposed CausalCF. Section \ref{sec:Eval} presents the experimental designs, results, and analysis. Finally, Section \ref{sec:Conclude} concludes the paper and discusses future work.
%%%%%%%%%%%%%%%%%%%%%%%%%%%%%%%%%%%%%%%%%%%%%%%%%%%%%%%%%%%%%%%%%%%%%%%%

\section{Background and Related Work}
\label{sec:background}
Many existing approaches in Causal RL assume that the SCM is known or partially known, making it impractical for many real-world problems. Existing Causal RL solutions are usually \emph{incomplete} and do not use counterfactuals. Counterfactuals enable the agent to learn the complete structural causal model and improve the robustness of the agent. In this section, we introduce the field of causality, review related work in causal RL, and discuss the approaches: Causal Curiosity and CoPhy.
\subsection{Causality}
The major limitation of current RL and DL methods is their lack of ability to generalize to out-of-distribution data. Combining causality with DRL can help solve the limitation. The \emph{SCM M} \cite{bar:etal2020} is a model that describes all the causal factors and relationships of a system and can be used to represent a specific RL environment. The agent learns a representation of the \emph{SCM M} which could be in the form of a causal graph. The aim is that the causal graph captures the invariances of the \emph{SCM M} that could be transferred to different environments or tasks with similar underlying causal structures. The terms causal modeling, causal reasoning, and causal inference all refer to the study of causality \cite{pearl2000models,halpern2020causes}. Different papers use the terms in subtly different ways. In this paper, the term causal inference will refer to the problems of inferring the SCM \cite{pearl2000models} from data and estimating the causal effects from a learnt representation of the SCM.

The agent can obtain different amounts of information about the \emph{SCM M} through the three different types of interactions. The three types of interactions are observation or association (seeing), intervention (doing or experimenting), and counterfactual (imagining and introspection). The Pearl causal hierarchy (PCH) \cite{bar:etal2020} categorizes the three types of interactions into different levels. In level 1 (association), the agent can detect regularities from passive observations. In level 2 (interventions), the agent can alter the environment and observe and predict its effects. In level 3 (counterfactuals), the agent can imagine alternate outcomes from unseen interventions or contexts. Higher-level interactions provide more information about the \emph{SCM M}. The causal hierarchy theorem (CHT) \cite{bar:etal2020} states that the three levels of causality cannot collapse and remain distinct. The CHT implies that agents who learn using lower-level interventions cannot make higher-level inferences like counterfactuals. The agent needs to use level 3 counterfactuals to learn the structural causal model. Counterfactuals require the agent to make inferences on new situations based on the past experiences of the agent. Therefore, counterfactuals prompt the agent to learn the invariant causal mechanisms across tasks or experiences. The ability of RL agents to perform counterfactuals improves its robustness.
%%%%%%%%%%%%%%%%%%%%%%%%%%%%%%%%%%%%%%%%
% \begin{figure}[H]
%       \centering
%       \includegraphics[scale=0.4]{Background_figs/PCH_levels.jpg}
%       \caption{Interactions categorised by the PCH, adapted from \cite{bar:etal2020}. Lower level interactions under determines higher level interactions.}
%       \label{fig:PCH_levels}
% \end{figure}
%%%%%%%%%%%%%%%%%%%%%%%%%%%%%%%%%%%%%%%%
\subsection{Related Work}
Currently, there is limited research on Causal RL. Existing approaches have limitations, such as assuming the SCM is known upfront, or partially known \cite{Madumal2020ExplainableRL,Ke2019LearningNC,dazeley2021explainable}, they are usually not complete  \cite{Sontakke2021CausalCR,Ahmed2021CausalWorldAR,Jin2020OfflineLO, Zhu2020CausalDW}, and they have not been applied to complex tasks like robotics \cite{Dasgupta2019CausalRF,Baradel2020COPHYCL}.

For instance, \cite{dazeley2021explainable} presents different approaches for achieving explainable robotics and mentions causal RL as one of the approaches. The paper \cite{dazeley2021explainable} also highlighted that one of the significant limitations for applying causal RL to real-world problems is that most existing approaches assume a known SCM. For example, \cite{Ke2019LearningNC} is a causal learning solution that uses observational data and interventions to learn the SCM of a Bayesian network. The approach assumes a partially known SCM and does not use counterfactuals. Additionally, \cite{Madumal2020ExplainableRL} assumes a known \emph{causal model} (causal relationships) and learns the causal factors through RL. In \cite{Madumal2020ExplainableRL}, valuable methods for evaluating the explainability of RL were provided. The generation and evaluation of explanations are out of the scope of this paper.

Some methods do not assume a known or partially known SCM but are incomplete Causal RL solutions. CausalWorld \cite{Ahmed2021CausalWorldAR} provides an environment where RL baselines can train using observations and interventions. However, the RL baselines do not use counterfactuals and causal representations. CausalWorld mainly provides a benchmark for evaluating causal RL solutions. Causal Curiosity \cite{Sontakke2021CausalCR} uses interventions and an abstract causal representation. However, Causal Curiosity does not use counterfactuals. Another approach \cite{Jin2020OfflineLO} makes use of human interventions and learns counterfactual predictions offline for real-world robotic grasping and manipulation tasks. A causal representation is not used in \cite{Jin2020OfflineLO}. Additionally, \cite{Zhu2020CausalDW} only uses observational data to learn a causal graph, and the solution struggles when graphs reach 50 nodes.

Methods that perform causal structure learning and are complete Causal RL solutions are not applied to more complex tasks like robotics. In \cite{Dasgupta2019CausalRF}, all three types of causal interactions were used to learn the causal structure of Bayesian networks, and it used a causal representation. The work demonstrated that all three types of interactions increased the performance of the RL agent. Moreover, CoPhy \cite{Baradel2020COPHYCL} is a causal learning solution that uses all three types of causal interactions and a causal representation to learn the physical dynamics of different objects in a simulated environment.

Causal Curiosity \cite{Sontakke2021CausalCR} and Cophy \cite{Baradel2020COPHYCL} are the most relevant to our chosen application area. We expand on them more.

\subsection{Causal Curiosity}
Causal Curiosity \cite{Sontakke2021CausalCR} is a Causal RL method that has been applied in CausalWorld. Causal Curiosity learns about the SCM from scratch using interventions and has two main training phases. The recursive training phase aims to learn a Cross Entropy Method optimized Model Predictive Control Planner (CEM MPC) \cite{Sontakke2021CausalCR} where K actions can be sampled. Each action provides information about one causal factor. This goal is achieved through maximizing the Curiosity reward. In the inference phase, K actions are sampled from the trained CEM MPC and applied to the RL training environment. The resulting observation is clustered and forms the causal representation (causalRep). The causalRep is then concatenated to all the observations to train the RL agent. The RL agent maximizes the CausalWorld reward.

Causal Curiosity uses interventions and provides an approach for training the RL agent with an abstract causal representation. However, the main drawback is that Causal Curiosity does not use counterfactuals. Counterfactuals are required for the agent to learn the complete underlying SCM and helps to improve the agent's robustness.

\subsection{Counterfactual Learning of Physics Dynamics (CoPhy)}
\begin{figure*}[h]
      \centering
      \includegraphics[scale=0.47]{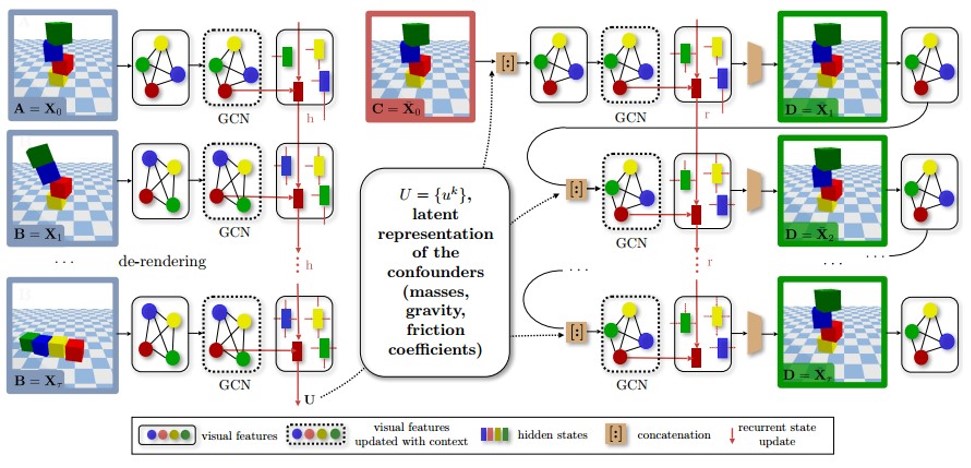}
      \caption{Overall architecture of the CoPhy solution \cite{Baradel2020COPHYCL}.}
      \label{fig:Cophy_original_pic}
\end{figure*}

CoPhy \cite{Baradel2020COPHYCL} provides a mechanism for performing counterfactuals in a supervised learning application to learn the physical dynamics of different objects in a simulated 3D environment. CoPhy is applied to predict the position, orientation, and stability of a given configuration of blocks for multiple time steps into the future. Figure \ref{fig:Cophy_original_pic} presents the overall architecture used. Observations A and B for $\tau$ time steps are used for training the entire model from end to end. The latent representation (confounders \emph{U}) or the causal representation is the embedding in the last layer of the recurrent neural network (RNN). An intervention is made on observation A (the initial state or $X_0$), and the result is observation C. Counterfactual predictions are made for $\tau - 1$ time steps as the initial intervened observation C is given. The de-rendering blocks in Figure \ref{fig:Cophy_original_pic} are used to convert the pixel observations into latent representations. The graph convolutional network blocks (GCN) add contextual information to the latent representations. Finally, the RNN temporally integrates the GCN blocks. CoPhy aims to minimize the mean squared error between the predictions and the actual states of the blocks. CoPhy learns about the underlying SCM from scratch using the three interactions in the PCH \cite{bar:etal2020}. As shown in Figure \ref{fig:Cophy_original_pic}, the causal representation (confounders \emph{U}) is used to make the counterfactual inferences for every time step. This approach allows us to train confounders \emph{U} \cite{Baradel2020COPHYCL} that capture the invariant causal mechanisms across the multiple time steps.

CoPhy is trained using supervised learning and requires many labelled images. DL methods cannot interact with the environment, and the interventions are usually handpicked. On the other hand, RL agents can interact with the environment and perform interventions based on their goals. Our proposed approach, CausalCF, adapts CoPhy for RL and enables RL agents to perform counterfactuals. To the best of our knowledge, this is the first time CoPhy has been adapted for use in RL.
%%%%%%%%%%%%%%%%%%%%%%%%%%%%%%%%%%%%%%%%
%%%%%%%%%%%%%%%%%%%%%%%%%%%%%%%%%%%%%%%%

%%%%%%%%%%%%%%%%%%%%%%%%%%%%%%%%%%%%%%%%%%%%%%%%%%%%%%%%%%%%%%%%%%%%%%%%

\section{Evaluation Environment}
\label{sec:causalworld}
CausalWorld \cite{Ahmed2021CausalWorldAR} is a benchmark environment created to propel causal and transfer learning research for robotics. CausalWorld provides a simulation environment (powered by PyBullet) for robotic manipulation tasks based on the open-source TriFinger robot. Sim2real transfer \cite{9308468} is easy to perform as CausalWorld is a realistic simulation of TriFinger and real-world physics. All the tasks in CausalWorld aim to arrange the blocks into their desired goal shape.
%\subsubsection{Observation space}

\begin{figure}[h]
      \centering
      \includegraphics[scale=0.57]{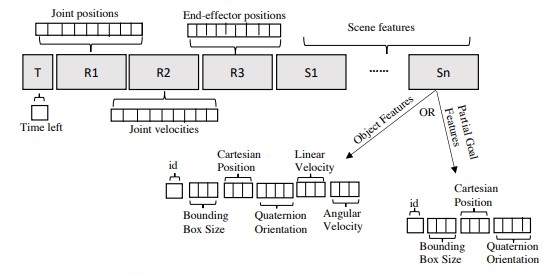}
      \caption{Structured observation \cite{Ahmed2021CausalWorldAR}.}
      \label{fig_design:CW_struct_observations}
\end{figure}

CausalWorld provides a structured observation type that presents the observation as a feature vector, as shown in Figure \ref{fig_design:CW_struct_observations}. We use structured observation to train CausalCF. Structured observation makes experimenting with the feasibility of the solution easier.

%\subsubsection{Action space}
The robot can be controlled through three different modes. The modes are joint positions, joint torques, and end-effector positions as indicated with the R1, R2, and R3 parameters in Figure \ref{fig_design:CW_struct_observations}. The action is specified for each of the three fingers on the robot. CausalWorld converts the action specified by the vector into motor actions that control the robot. There are constraints set on the values of the action parameters based on the physical properties of the robot. In this work, the RL agent uses joint positions to control the robot. End-effector positions are non-deterministic for the current version of CausalWorld, and joint positions are more intuitive to use than joint torques.
%\subsubsection{Rewards}

The consecutive actions of the Causal RL agent are chosen to maximize the cumulative rewards in CausalWorld. Rewards are defined differently for different tasks. Rewards are a sum of different objective functions, and each function aims to affect the agent's behavior in a particular way. The user can specify the weights of each function. RL agents would then prioritize to maximize the functions with larger weights. For example, the first objective of the agent in any task is to get its fingers to the blocks, and this objective function would have the largest weight. The weights used for the tasks in this paper are the same as those used in the original CausalWorld paper \cite{Ahmed2021CausalWorldAR}.

CausalWorld will be used for the evaluation of CausalCF. CausalWorld provides an evaluation pipeline that consists of 12 protocols as described in Table \ref{tab_eval:protocols} \cite{Ahmed2021CausalWorldAR}. Each of the 12 protocols applied in the evaluation pipeline modifies a specific subset of the CausalWorld variables and defines a range of values that the variables can take \cite{Ahmed2021CausalWorldAR}. Space A and B \cite{Ahmed2021CausalWorldAR} in Table \ref{tab_eval:protocols} define different ranges of values the CausalWorld variables can take. Agents are evaluated over 200 episodes in each of the 12 protocols.
\begin{table*}[h]
\caption{Table is adapted from CausalWorld \cite{Ahmed2021CausalWorldAR}. Variables: bp - block pose, bm - block mass, bs - block size, gp - goal pose and ff - floor friction.}
\label{tab_eval:protocols}
\centering
\begin{tabular}{| p{1.16cm} | p{0.6cm} | p{0.6cm} | p{0.6cm} | p{0.6cm} | p{0.6cm} | p{0.6cm} | p{0.7cm} | p{1cm} | p{1cm} | p{1cm} | p{0.8cm} | p{0.8cm} |}
\hline
\textbf{} & \textbf{P0} & \textbf{P1} & \textbf{P2} & \textbf{P3} & \textbf{P4} & \textbf{P5} & \textbf{P6} & \textbf{P7} & \textbf{P8} & \textbf{P9} & \textbf{P10} & \textbf{P11}\\
\hline
\textbf{Space} & A & A & B & A & A & A & B & A & B & B & A & B\\
\hline
\textbf{Var} & - & bm & bm & bs & bp & gp & bp, gp & bp, gp, bm & bp, gp, bm & bp, gp, bm, ff & all & all\\
\hline
\end{tabular}
\end{table*}

%%%%%%%%%%%%%%%%%%%%%%%%%%%%%%%%%%%%%%%%
%%%%%%%%%%%%%%%%%%%%%%%%%%%%%%%%%%%%%%%%

%%%%%%%%%%%%%%%%%%%%%%%%%%%%%%%%%%%%%%%%%%%%%%%%%%%%%%%%%%%%%%%%%%%%%%%%

\section{Design of CausalCF}
\label{sec:causalCF}
CausalCF has two main training phases, \emph{counterfactual training} and \emph{agent training}, as shown in Figure \ref{fig_design:CausalCF_overview}. Counterfactual training produces the causal representation used to train the agent in agent training. An additional component, \emph{iteration training}, allows the agent to update its causal representation.

\subsection{Counterfactual Training}
\begin{figure*}[h]
\begin{subfigure}{\textwidth}
    \centering
    \includegraphics[scale=0.55]{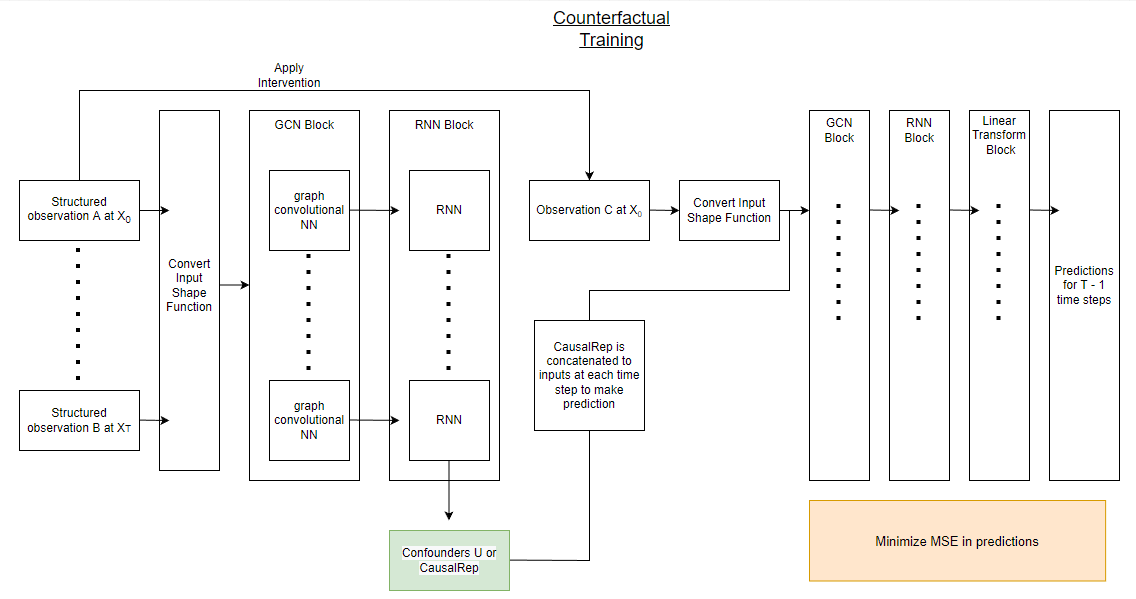}
    \caption{Counterfactual training. This modified version of CoPhy is called Counterfactual model (CF model).}
    \label{subfig_design:CausalCF_CFtrain}
\end{subfigure}
\begin{subfigure}{\textwidth}
    \centering
    \includegraphics[scale=0.5]{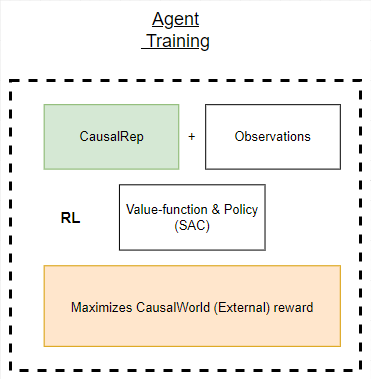}
    \caption{Agent training}
    \label{subfig_design:CausalCF_AgentTrain}
\end{subfigure}
\caption{Overview design of the CausalCF solution. CausalCF has two phases of training like in Causal Curiosity. The diagram in the Counterfactual training is adapted from CoPhy \cite{Baradel2020COPHYCL}.}
\label{fig_design:CausalCF_overview}
\end{figure*}

The aim of counterfactual training (Figure \ref{subfig_design:CausalCF_CFtrain}) is to obtain a useful causal representation that can be concatenated to the observations to train the RL agent. For this purpose, we adapt CoPhy and train it from end to end in counterfactual training.

The original CoPhy solution had a derendering block \cite{Baradel2020COPHYCL} that converted pixel images into latent representations before passing it to the GCN Block (Figure \ref{fig:Cophy_original_pic}). CoPhy is adaptable to the number of time steps and objects. In contrast to CoPhy, CF model (Figure \ref{subfig_design:CausalCF_CFtrain}) receives structured observation (Figure \ref{fig_design:CW_struct_observations}) instead of images. Therefore, we replaced the derendering block and dependent processes with new functions. We replaced the derendering block with the \emph{convert input shape} function (Algorithm \ref{alg:convert_input}), which converts the structured observation into a representation adaptable to a varying number of time steps and objects. The representation has the shape (T, K, 56), where T is the number of timesteps, K is the number of objects, and a feature vector of length 56 (Parameters T+R1+R2+R3+Sn in Figure \ref{fig_design:CW_struct_observations}) is enough to describe the state of the robot and any single object. We made modifications throughout the entire architecture to process the structured observations. We used the RL agent to generate the observations and ground truths for training the CF model. Algorithm \ref{alg:train_cf_model} describes the process of generating the training data and training CF Model. CF Model makes counterfactual predictions about the future states of the environment from the effect of interventions. CF Model aims to minimize the mean squared error between the counterfactual predictions and the actual states of the environment. 

The ability to perform counterfactual predictions provides a mechanism for improving the explainability of the RL agent. Causal-CF could provide local explanations \cite{Puiutta2020ExplainableRL} for a particular action or decision by performing counterfactuals and imagining the future consequences of different actions based on the causal knowledge of the agent. Explanations involve providing the different outcomes with the corresponding rewards as RL agents aim to maximize cumulative rewards. The probability of success \cite{Cruz2021ExplainableRS} could be used instead of rewards because rewards vary from task to task. If the imagined outcomes and probability of success match the knowledge of the human expert, then the decisions or actions of the agent can be trusted. Generation and evaluation of counterfactual explanations are out of the scope of this paper, and we focus on utilizing counterfactuals to improve performance and robustness.
%%%%%%%%%%%%%%%%%%%%%%%%%%%%%%%%%%%%%%
\begin{algorithm}[ht]
\caption{Algorithm for component Convert Input Shape}\label{alg:convert_input}
$num\_timesteps \gets T$\;
$num\_objects \gets K$\;
$desired\_input\_obs \gets [T, K, 56]$ \tcp*[l]{Empty array for storing new input}
$lens\_obs \gets stack\_obs.shape[1]$ \tcp*[l]{Length of the Structured observation}
\tcp*[l]{stack\_obs represent the Struct\_obs\_ab and Struct\_obs\_c in Figure \ref{subfig_design:CausalCF_CFtrain}}
\For{$t = 0, num\_timesteps$}{
  \For{$k = 0, num\_objects$} {
    $desired\_input\_obs[t, k, :28] \gets stack\_obs[t, :28]$ \tcp*[l]{Store T, R1, R2 and R3}
    $obj\_index\_low \gets 28 + (k * 17)$\;
    $obj\_index\_up \gets obj\_index\_low + 17$\;
    $desired\_input\_obs[t, k, 28:45] \gets stack\_obs[t, obj\_index\_low:obj\_index\_up]$ \tcp*[l]{Store CW Object features}
    $part\_goal\_index\_low \gets (28 + (num\_objects * 17)) + (k * 11)$\;
    $part\_goal\_index\_up \gets part\_goal\_index\_low + 11$\;
    \tcp*[l]{Store Partial goal features}
    \If{$part\_goal\_index\_up \geq len\_obs$}{
        $desired\_input\_obs[t, k, 45:] \gets stack\_obs[t, part\_goal\_index\_low:]$\;
    }
    \Else{
        $desired\_input\_obs[t, k, 45:] \gets stack\_obs[t, part\_goal\_index\_low:part\_goal\_index\_up]$\;
    }
  }
}
\end{algorithm}
%%%%%%%%%%%%%%%%%%%%%%%%%%%%%%%%%%%%%%
%%%%%%%%%%%%%%%%%%%%%%%%%%%%%%%%%%%%%%
\begin{algorithm}
\caption{Algorithm shows how data is generated to train CF\_model}\label{alg:train_cf_model}
$max\_iter \gets 40$\;
$num\_timesteps \gets 30$\;
$initial\_obs$ \tcp*[l]{Stores initial Structured observation when environment resets}
\For{curr\_iter = 0, max\_iter}{
    $action \gets RL\_agent.act(initial\_obs)$\;
    \For{t = 0, num\_timesteps}{
        $next\_obs \gets env.step(action)$\;
        $action \gets RL\_agent.act(next\_obs)$\;
        $stack\_input\_obs \gets stack\_input\_obs.concatenate(next\_obs)$\;
    }
    $desired\_input\_obs \gets Convert\_input\_shape(stack\_input\_obs)$ \tcp*[l]{Calls Convert Input Shape component of CF\_model}
    $pose\_3d\_ab \gets process1(desired\_input\_obs)$\ \tcp*[l]{Turns numpy array to a torch tensor for using GPU}
    $observation\_c \gets goal\_intervention(env)$\;
    $desired\_obs\_c \gets Convert\_input\_shape(observation\_c)$\;
    $pose\_3d\_c \gets process2(desired\_obs\_c)$\ \tcp*[l]{Turns numpy array to a torch tensor for using GPU}
    $pred\_D\_out, pred\_stability, causal\_rep \gets CF\_model(pose\_3d\_ab, pose\_3d\_c)$ \tcp*[l]{Pass inputs through all the other components of CF\_model}
    $action \gets RL\_agent.act(observation\_c)$ \tcp*[l]{Generate ground-truth for CF\_model predictions}
    \For{t = 0, num\_timesteps - 1}{
        $next\_obs \gets env.step(action)$\;
        $action \gets RL\_agent.act(next\_obs)$\;
        $stack\_input\_obs \gets stack\_input\_obs.concatenate(next\_obs)$\;
    }
    $desired\_input\_obs \gets Convert\_input\_shape(stack\_input\_obs)$\;
    $actual\_D\_out \gets process3(desired\_input\_obs)$\ \tcp*[l]{Same purpose as process1 and process2}
    $mse\_3d \gets Calc\_loss(pred\_D\_out, actual\_D\_out)$\;
    Update CF\_model with loss calculation\\
    Reset environment
}
\end{algorithm}
%%%%%%%%%%%%%%%%%%%%%%%%%%%%%%%%%%%%%%
\subsubsection{Train CF model with Pretrained agent}
CF model was trained with a pretrained RL agent provided by CausalWorld. The pretrained agent can perform useful interventions or actions that allow the causal representation to capture causal factors. The causal representation is concatenated to all the observations for training the RL agent. Using the pretrained agent enables the viability of the CF model to be evaluated in the early stages, and the total training time for CausalCF is reduced. CF model is trained for 15 epochs, and the model is trained for 40 iterations for each epoch. In each epoch, an intervention is made on the goal shape and the block mass. The pretrained agent generates observations for 30 timesteps to train the CF model from end to end in each iteration (Algorithm \ref{alg:train_cf_model}). Approaches without using pretrained agents involve training the RL agent with an initialized causal representation. When the agent converges or reaches a desirable performance, the agent can iterate between counterfactual training and agent training to update the causal representation.

\subsection{Agent Training}
\label{subsec:Agent_train}
The resulting causal representation from counterfactual training is concatenated to all the observations for training the RL agent in agent training. The RL agent aims to maximize the CausalWorld reward in agent training (Figure \ref{subfig_design:CausalCF_AgentTrain}). The idea of using two training phases and concatenating the causal representation to all the observations used for training the agent is inspired by Causal Curiosity. Interventions on the goal shape (pose) are performed every episode in agent training. CausalCF has an additional component where it iterates between counterfactual and agent training.

\subsection{Iteration Training}
The RL agent iterates between counterfactual and agent training after it has trained for 1.5 million time steps in agent training. Iteration between counterfactual and agent training happens every 500,000 time steps. The agent is trained for a total of 7 million time steps. The iteration between the two training phases allows the agent to update the causal representation or its causal knowledge the better the agent gets. The additional iterations for counterfactual training will enable the agent to capture more information about the SCM.

%%%%%%%%%%%%%%%%%%%%%%%%%%%%%%%%%%%%%%%%
%%%%%%%%%%%%%%%%%%%%%%%%%%%%%%%%%%%%%%%%

\section{Evaluation}
\label{sec:Eval}
In this section, we present the evaluation of CausalCF in CausalWorld. We describe the metrics, training parameters, and stages of evaluation for CausalCF. We also discuss the performance of CausalCF in the stages of evaluation. 
\subsection{Metrics and Parameters}
In CausalWorld, fractional success is often used to measure agent performance. Fractional success is calculated as:
\begin{equation}
    fractional\_success = \frac{\sum^{n}_{i=1}intersection(O\_vol_i, G\_vol_i)}{\sum^{n}_{i=1}G\_vol_i} 
\label{equation:frac_success}
\end{equation}
Where n represents the number of objects, $O\_vol_i$ represents the current pose of the i-th object, and $G\_vol_i$ represents the desired goal pose of the i-th object. The function $intersection(O\_vol_i, G\_vol_i)$ calculates the overlapping volume between the current block pose and the objects' goal block pose. A higher fractional success will correspond to more precise object manipulations. CausalCF uses fractional success for all the evaluations.

We use Soft-Actor Critic (SAC) \cite{haarnoja2018soft} for all the experiments in this work and parameters are given in Table \ref{tab_design:sac_config}. Table \ref{tab_eval:train_param} shows the training parameters used for all tasks in which CausalCF is evaluated.
\begin{table}[H]
\centering
\caption{Parameter configuration for the SAC algorithm.}
\begin{tabular}{rll}\toprule
%\hline
\textbf{Parameters} & \textbf{Value}\\
\midrule
\textbf{gamma} & 0.95 \\
\textbf{tau} & 1e-3 \\
\textbf{ent\_coef} & 1e-3 \\
\textbf{target\_entropy} & auto \\
\textbf{learning\_rate} & 1e-4 \\
\textbf{buffer\_size} & 1 000 000 \\
\textbf{learning\_starts} & 1000 \\
\textbf{batch\_size} & 256 \\
\bottomrule
\end{tabular}
\label{tab_design:sac_config}
\end{table}

\begin{table}[H]
\caption{Training parameters for each task used to evaluate CausalCF.}
\begin{tabular}{rll}\toprule
%\hline
\textbf{} & \textbf{Pushing \& Picking}\\
\midrule
\textbf{Total time steps} & 7000000\\
\textbf{Training time} & $\approx$22 hours\\
\textbf{Episode length} & 834\\
\textbf{Number of episodes} & 8393\\
\textbf{Skipframe} & 3\\
\textbf{Space} & A\\
\textbf{Checkpoint frequency} & 500000\\
\bottomrule
\end{tabular}
\label{tab_eval:train_param}
\end{table}

\subsection{Evaluation Stages}
There are two different types of evaluation performed on CausalCF: \emph{component testing} and \emph{causalRep transfer}.

Component testing evaluates the contributions of the different components of the CausalCF design to the robustness of the RL agent. We compare \emph{CausalCF (iter)} (CausalCF) with 3 baselines: \emph{Counterfactual + Intervene} (CausalCF without iterating between Counterfactual and Agent training), \emph{Intervene} (uses interventions but no counterfactuals), and \emph{no\_Intervene} (uses no interventions and no counterfactuals, the SAC baseline). The different RL models are passed through an evaluation pipeline that CausalWorld provides to measure generalizability.

We demonstrate the transferability of the causal representation across different tasks (and therefore increased robustness) in causalRep transfer. The causal representation learnt by CausalCF in the Pushing task is transferred to train a Causal RL agent in the Picking task. The performance of \emph{transfer Causal\_rep + Intervene} (agent that used the transferred causal representation) is compared against \emph{Intervene} (agent does not use a causal representation).

The training environment for \emph{no\_Intervene} is the same as P0 (Table \ref{tab_eval:protocols}), where no interventions on the CausalWorld variables are used. The training environment for all the other baselines is the same as P5, where interventions were made on the goal pose (Section \ref{subsec:Agent_train}) in Space A (Table \ref{tab_eval:train_param}). All the other protocols will measure the out-of-distribution robustness of the baselines. For example, P6 modifies the objects' block pose, and goal pose with the ranges of values provided in Space B (Section \ref{sec:causalworld}). The result of the interventions in P6 is that the agents will have to perform new grasps on the block and encounter new goal poses that are not seen in training.

\subsection{Component Testing}
\begin{figure}[htbp]
\begin{subfigure}{\columnwidth}
      \centering
      \includegraphics[scale=0.49]{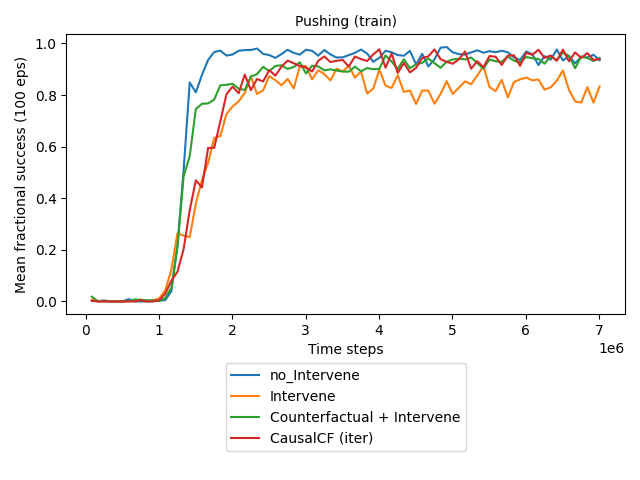}
      \caption{Training performance. The x-axis is the training time steps, and the y-axis is the mean of the fractional successes for every 100 episodes.}
      \label{fig_eval:plot_push_train}
\end{subfigure}
\begin{subfigure}{\columnwidth}
      \centering
      \includegraphics[scale=0.49]{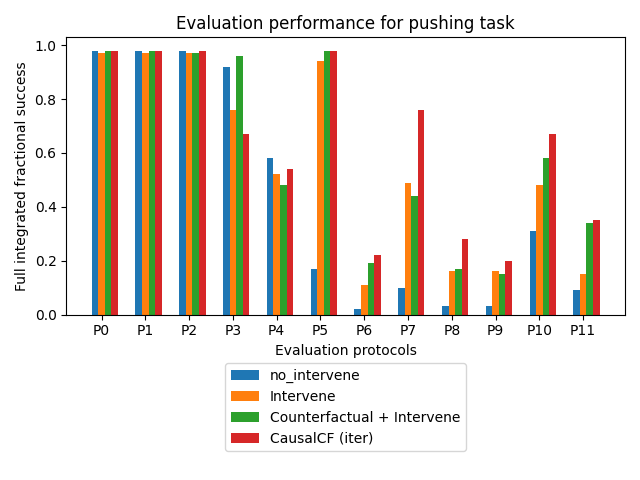}
      \caption{Evaluation performance. The x-axis shows the 12 protocols (Table \ref{tab_eval:protocols}), and the y-axis is the full integrated fractional success \cite{Ahmed2021CausalWorldAR} which is the mean fractional success over all 200 episodes for each evaluation protocol.}
      \label{fig_eval:plot_push_eval}
\end{subfigure}
\caption{Results for Component Testing in the Pushing task.}
\label{fig_eval:Push_task}
\end{figure}

The training performance of the different RL solutions in the Pushing task is provided in Figure \ref{fig_eval:plot_push_train}. Some interventions might lead to unsolvable scenarios, and the agent will receive a low fractional success for these rare scenarios. These scenarios can happen in the later stages of training. The mean fractional success is used to produce a smoother plot of the training performance.

No interventions are used to train the \emph{no\_Intervene} agent. Therefore, the environment for the \emph{no\_Intervene} agent is more repetitive, and the agent was able to converge to a good training performance faster than all the other solutions. The \emph{no\_Intervene} agent converged to a fractional success of higher than 0.9 in about 1.5 million time steps (Figure \ref{fig_eval:plot_push_train}). In contrast to the training environment used for \emph{no\_Intervene}, all the environments used for the other RL solutions intervened on the goal shape at every episode. Interventions make the tasks more challenging. The \emph{Intervene} solution could only converge to an approximate fractional success of 0.8 in 7 million time steps (Figure \ref{fig_eval:plot_push_train}). CausalCF outperformed \emph{Intervene} in the same training environment and converged to the same performance as the \emph{no\_Intervene} solution (Figure \ref{fig_eval:plot_push_train}). It took CausalCF longer to converge than \emph{no\_Intervene} because the training environment for CausalCF is more challenging. \emph{CausalCF (iter)} took longer to converge than \emph{Counterfactual + Intervene} (Figure \ref{fig_eval:plot_push_train}). The variations to the causal representation during the iteration require the RL agent to partially re-learn and converge more slowly. However, \emph{CausalCF (iter)} should be more robust than all the other solutions trained. The robustness of the RL solutions is measured through the evaluation pipeline provided by CausalWorld.

The evaluation pipeline provided by CausalWorld consists of 12 protocols as described in Table \ref{tab_eval:protocols} \cite{Ahmed2021CausalWorldAR}. The performance of the different RL solutions in the evaluation pipeline is provided in Figure \ref{fig_eval:plot_push_eval}. All the RL solutions performed better for protocols that used space A than B (Figure \ref{fig_eval:plot_push_eval}) because they were all trained in space A. Interventions on different causal variables create different challenges. All the solutions performed worse on the protocols that intervened on the block pose (Figure \ref{fig_eval:plot_push_eval}). The change in block pose requires the agent to consider the initial grasp on the block, which the agent did not consider as much in training. \emph{Intervene} generalized better than \emph{no\_Intervene} as interventions on one or a subset of causal variables are better than no intervention \cite{lee2018structural,CausalRLElias}. CausalCF was more robust than all the other solutions as the agents used counterfactuals to capture more causal information about the underlying SCM. \emph{CausalCF (iter)} performed equal to or better than the other solutions in 10 out of the 12 protocols and is the most robust (Figure \ref{fig_eval:plot_push_eval}). \emph{CausalCF (iter)} allows the agent to update its causal representation further as the agent gets better.
%%%%%%%%%%%%%%%%%%%%%%%%%%%%%%%%%%%%%%%%
%%%%%%%%%%%%%%%%%%%%%%%%%%%%%%%%%%%%%%%%
\subsection{CausalRep Transfer}
\begin{figure}[htbp]
\begin{subfigure}{\columnwidth}
      \centering
      \includegraphics[scale=0.43]{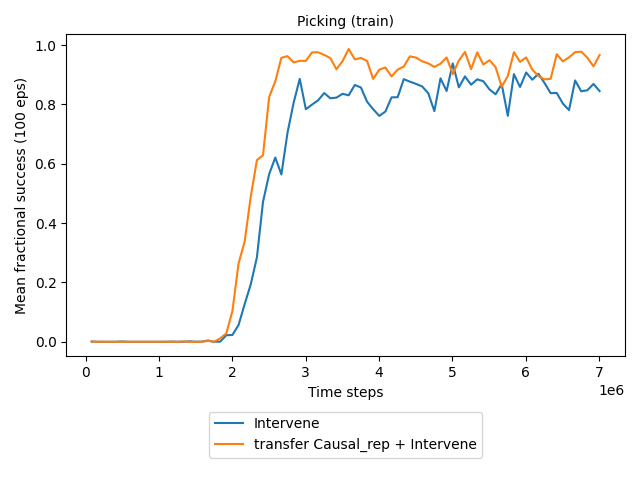}
      \caption{Training performance}
      \label{fig_eval:plot_pick_train}
\end{subfigure}
\begin{subfigure}{\columnwidth}
      \centering
      \includegraphics[scale=0.41]{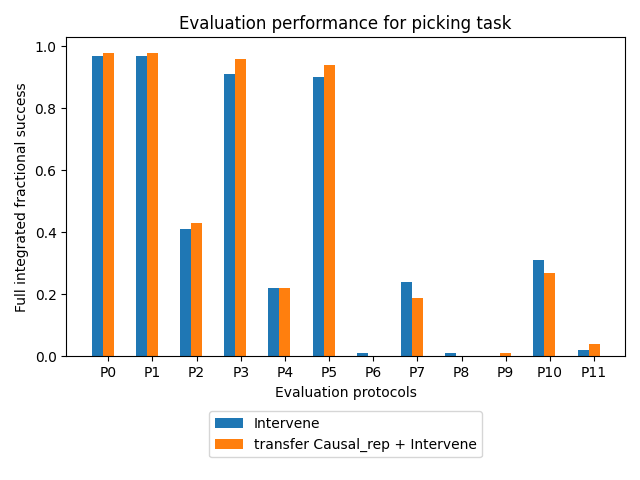}
      \caption{Evaluation performance}
      \label{fig_eval:plot_pick_eval}
\end{subfigure}
\caption{Results for causalRep transfer in the Picking task. The 12 protocols applied are the same as it is described in Table \ref{tab_eval:protocols}}
\label{fig_eval:Pick_task}
\end{figure}

The tasks in CausalWorld have similar causal structures. The learnt causal representation of the \emph{CausalCF (iter)} solution is transferred from the Pushing task to train a new RL agent in the Picking task. Two solutions are compared, and the training performance is shown in Figure \ref{fig_eval:plot_pick_train}. Both solutions used the same interventions. Intervention is made on the goal shape of the block at every episode. The \emph{transfer Causal\_rep + Intervene} solution converged to a fractional success of higher than 0.9 (Figure \ref{fig_eval:plot_pick_train}). The \emph{Intervene} solution converged to a fractional success of about 0.8 (Figure \ref{fig_eval:plot_pick_train}). \emph{transfer Causal\_rep + Intervene} converged to a better training performance than \emph{Intervene}. The causal representation provides additional prior causal knowledge for the RL agent.

Figure \ref{fig_eval:plot_pick_eval} shows the performance of the two RL solutions in the evaluation pipeline. \emph{transfer Causal\_rep + Intervene} performed equal to or better than \emph{Intervene} in 8 out of the 12 evaluation protocols (Figure \ref{fig_eval:plot_pick_eval}). The causal representation learnt from the Pushing task helps the agent perform better and become more robust in the Picking task. This result confirms that the causal representation captures causal mechanisms through interventions and counterfactuals. \emph{transfer Causal\_rep + Intervene} could have achieved better evaluation performance if the agent could update the causal representation with new causal information in the Picking task.
%%%%%%%%%%%%%%%%%%%%%
%%%%%%%%%%%%%%%%%%%%%%%%%%%%%%%%%%%%%%%%

%%%%%%%%%%%%%%%%%%%%%%%%%%%%%%%%%%%%%%%%%%%%%%%%%%%%%%%%%%%%%%%%%%%%%%%%
\section{Conclusions and Future Work}
\label{sec:Conclude}
In this work, we have presented CausalCF, the first complete Causal RL solution that successfully tackles complex RL robotic tasks. We have shown that CausalCF improves the robustness of RL and that none of the design components of CausalCF are redundant in component testing. The results from causalRep transfer show that the causal representation captures the causal mechanisms that remain invariant across tasks. CausalCF improves the robustness of RL and provides mechanisms to help improve the explainability of RL. Future work includes Multitask RL, where the skills obtained from multiple simpler tasks are applied to more challenging tasks. We also hope to evaluate CausalCF in sim2real transfer (mentioned in Section \ref{sec:causalworld}).
%%%%%%%%%%%%%%%%%%%%%%%%%%%%%%%%%%%%%%%%%%%%%%%%%%%%%%%%%%%%%%%%%%%%%%%%

%\section*{Acknowledgments}
\begin{acks}
This work was funded in part by the SFI-NSFC Partnership Programme Grant Number 17/NSFC/5224 and Science Foundation Ireland Grant number 18/CRT/6223. 
\end{acks}
%%% The next two lines define, first, the bibliography style to be 
%%% applied, and, second, the bibliography file to be used.
\clearpage
\bibliographystyle{ACM-Reference-Format} 
\bibliography{sample}

%%%%%%%%%%%%%%%%%%%%%%%%%%%%%%%%%%%%%%%%%%%%%%%%%%%%%%%%%%%%%%%%%%%%%%%%

\end{document}